\documentclass[runningheads]{llncs}
\usepackage[T1]{fontenc}
\usepackage{graphicx}
\usepackage{amsmath}
\usepackage{amssymb}
\usepackage{booktabs}
\usepackage{algorithm}
\usepackage{algpseudocode}
\usepackage[demo,abs]{overpic}

\def\bU{{\bf U}}

\def\bl{{\bf l}}

\def\br{{\bf r}}

\def\bv{{\bf V}}

\def\bt{{\bf t}}
\def\bw{{\bf w}}
\def\bPi{{\Pi}}
\def\b0{{\bf 0}}

\def\br{{\bf r}}
\begin{document}
\title{Robust and Accurate Cylinder Triangulation}
%

\author{Anna Gummeson \and Magnus Oskarsson}
%
%

\institute{Centre for Mathematical Sciences, Lund University, Sweden \\
\email{\{anna.gummeson, magnus.oskarsson\}@math.lth.se}}
\maketitle              
\begin{abstract}
In this paper we present methods for triangulation of infinite cylinders from image line silhouettes. We show numerically that   linear estimation of a general quadric surface is  inherently a badly posed problem. Instead we propose to constrain the conic section to a circle, and give algebraic constraints on the dual conic, that models this manifold. Using these constraints we derive a fast minimal solver based on three image silhouette lines, that can be used to bootstrap robust estimation schemes such as RANSAC. We also present a constrained least squares solver that can incorporate all available image lines for accurate estimation. The algorithms are tested on both synthetic and real data, where they are shown to give accurate results, compared to previous methods. 

\keywords{Reconstruction  \and Robust estimation \and Cylinders.}
\end{abstract}
\section{Introduction}
The world is a complex place. In order to make sense of it, we as humans typically make simplifications in order to understand and interpret our surroundings. The use of geometric primitives, to represent and build up the world, has been  advocated both from a computer vision perspective, via the influential  ideas of Marr \cite{marr1978representation}, and from a psychology perspective, via the recognition-by-components (RBC) theory of Biederman \cite{biederman1985human}, for a long time.  These ideas were put forward mostly in the context of object recognition. Today,  applications for extraction of semantic information from images and scenes are typically highly data driven, with representations largely learned and coded implicitly in neural network architectures.  Geometric estimation in Structure from Motion (SfM) and Simultaneous Localization and Mapping (SLAM) application are, on the other hand, often based on explicit representations (sparse 3D-points and camera matrices) even if the image feature representations are learned.  There are  today mature methods that do SfM and SLAM efficiently \cite{mur2015orb,schoenberger2016sfm,schoenberger2016mvs}. Sparse point clouds are well suited for matching, camera geometry estimation and optimization. However, they do not scale well for very large scenes, they are often not stable over time  \cite{sattler2018benchmarking}, and are bad for downstream tasks such as interpretation and recognition. For this reason we are interested in investigating the use of mid-level representations that carry more semantic meaning than simple points do. One such possible primitive shape is a cylinder. In this paper we will specifically study the problem of how one can efficiently estimate the 3D structure of a cylinder given views of it, in a number of images. We will show how one can triangulate the shape from silhouette lines in two or more images, in both a robust and an accurate way. Note that we in this paper restrict image information to the silhouettes of the sides of cylinders, and not the apparent contours of the cross sections, i.e.\ we consider infinitely long cylinders (analogous to using lines or line segments for line geoemetry). 
Working with lines in images is quite difficult, but line segment detectors have matured over the last years, \cite{xu2021line,dai2021fully,von2012lsd,pautrat2021sold2} and we foresee that they will soon be applicable for many general computer vision tasks. 

A cylinder is an example of a quadric surface, and can basically be estimated linearly, given silhouette lines in a number of views with known camera geometry. However, as we will show in a number of real examples, this problem is inherently ill-posed, and in order to find realistic solutions we need to constrain the estimation process. Our contributions in the paper are that we 
(i) formulate polynomial constraints on a quadric being a cylinder, 
(ii) derive a minimal solver based on these constraints that can be used for robust estimation, and
(iii)  derive a non-iterative constrained least squares solver that can be used for accurate estimation.  Code for solvers will be made publicly available upon acceptance.

There exists  previous work on a number of geometric problems involving quadrics and conics, but most work is concerned with general projective models, e.g. \cite{de1993conics,ma1996ellipsoid}.
For planar conics and circles there  are several results on pose, homography estimation and relative pose, e.g. in \cite{kaminski2003multiple,mudigonda2004geometric,frosio2012linear,mei2017monocular,frosio2016camera}. The two view relative pose problem was considered in \cite{kahl1998using,quan1996conic}  which is closely related to the theory for  general silhouettes, \cite{astrom-cipolla-etal-ijcv-99,astrom-kahl-itpam-99}.
In  \cite{navab2006canonical} the authors present  algorithms for pose, 3D reconstruction and structure from motion from cylinders.  These methods do not, however, consider robust or least square solutions for triangulation.
Work from \cite{winkler2013curve} and \cite{sun2019camera} are addressing camera calibration using cylinders, but using the apparent contours of the ends of finite cylinders instead. 
The inverse problem, camera position from known cylinders, is covered in \cite{gummesonengman2022fast}.
 The triangulation problem has also  been considered for other  geometric shapes, such as lines \cite{bartoli2005structure}, planar conics  \cite{josephson2008triangulation} and more complex surfaces based on line representations \cite{keppel1975approximating}.

\section{Cylinder geometry}
We will now describe some of the underlying geometric concepts  used to model projections of cylinders.   For a more detailed description of the projective aspects of quadrics and conics, we refer the reader to the classic book on projective geometry by Semple and Kneebone \cite{semple1979algebraic}. 

A surface defined by a quadratic expression in three dimensions, is known as a quadric. A point $\bU \in \mathbb{P}^3$ lies on a quadric if 
\begin{equation}
\bU ^TC\bU= 0,
\end{equation}
where the point $\bU$ is given in homogeneous coordinates, and $C$ is a $4\times4$ symmetric matrix defining the \emph{quadric locus}. If $C$ has full rank it is  a \emph{proper} quadric. A proper quadric can also directly be described using all planes that are tangent to the quadric surface, $\bPi \in \mathbb{P}^3$, 
\begin{equation}
\bPi ^TD\bPi= 0,
\label{eq:dualquadric}
\end{equation}
where again $\bPi$ is given in homogeneous coordinates, and $D$ is a symmetric $4\times4$ matrix now defining the \emph{quadric envelope}. Here $D$ is the dual representation of $C$ and given by $D \sim C^{-1}$,
where $\sim$ denotes equality up to a non-zero scalar.

\begin{figure}
	\centering
	\begin{overpic}[width=0.45\columnwidth]{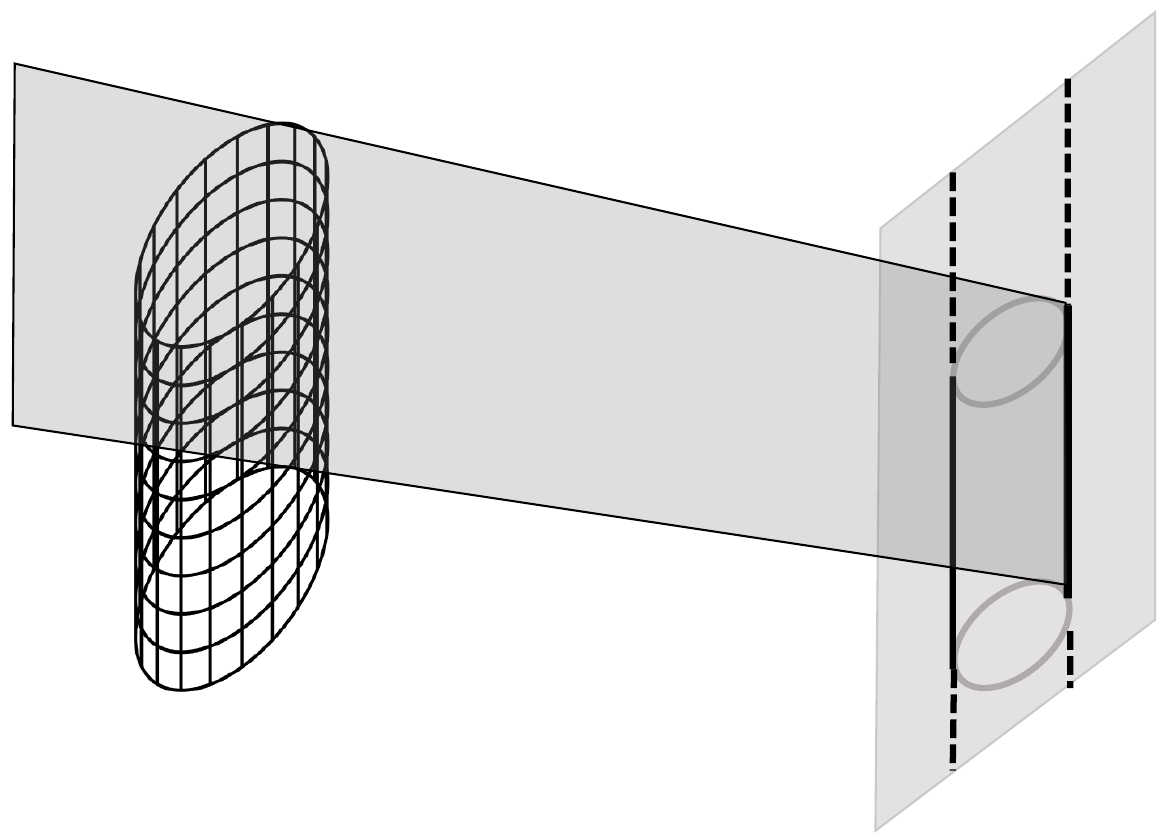}
		\put(63,65){$\Pi = P^T\bl$}
		\put(147,50){$\bl$}
		\put(25,10){$D$}
	\end{overpic}
	\caption{A cylinder is a special quadric surface, a cone with intersection point at infinity. The projection is defined by two silhouette lines.}
	\label{fig:quadricprojections}
\end{figure}

The projection of a quadric surface on a plane is a conic.  Using the direct representation leads to quite involved expressions, but the  projection can be described in a  simple way using the dual formulation. If the quadric is defined by (\ref{eq:dualquadric})
and the projected conic  by
\begin{equation}
\bl^Td\bl = 0,
\end{equation}
then, given a camera matrix $P$, we have
\begin{equation}
d \sim PDP^T.
\end{equation}
Note that in this expression the projection is linear in $D$.
If we have extracted a tangent line, $\bl$, to a conic in an image, then this directly gives one constraint on the camera geometry
\begin{equation}
0 = \lambda \bl^Td\bl = \bl^TPDP^T\bl.
\label{eq:quadricprojection}
\end{equation}
Note that this is not the same as an ordinary line constraint, since we don't have line-to-line correspondences.

When the quadric locus $C$ doesn't have full rank,  then $C$ has an eigenvalue which is zero. The corresponding eigenvector $\bv$ will then represent a point on the quadric surface since
\begin{equation}
\bv^TC\bv = \bv^T(C\bv) =   \bv^T \b0 = 0.
\end{equation}
Now, for any other point $\bU$ that lies on the quadric,
$\bv+\lambda \bU$ will also lie on the quadric. This means that all points on the quadric surface will lie on lines through $\bv$.  This defines an elliptic cone in space.  
All planes through $\bv$ will be tangent planes, however these will not uniquely define the quadric surface. By restricting the tangent planes to those that contain a line $\bv+\lambda \bU$, for some $\bU$ on the cone, we will get a unique definition of the surface. By cutting the cone with any plane (not tangent to the cone) we can define the dual representation $D$ of $C$, using the conic sections and envelopes in this plane. This means that we now can define the cone using two equations, 
\begin{equation}
\{ \Pi | \Pi^T \bv = 0, \Pi^T D \Pi = 0 \} ,
\label{eq:dualcone}
\end{equation}
where $\Pi$ are planes given in homogeneous coordinates.
The projection silhouette of a cone is simply two intersecting lines. Given a camera $P$, each of these lines, $\bl$, will fulfill two constraints, directly given by  \eqref{eq:dualcone},
\begin{align}
\bl^TP\bv & = 0,
\label{eq:coneprojection1}
\\
\bl^TPDP^T\bl &  = 0,
\label{eq:coneprojection2}
\end{align}
since the tangent plane is given by $\Pi = P^T\bl$.
 The intersection point $\bv$ of a cone can be any point in  $\mathbb{P}^3$. If this point is placed at infinity, the Euclidean interpretation in $\mathbb{R}^3$ is that the lines of the cone will be parallel. This means that the cone in this case will be a cylinder. 
A point $\bv$ at infinity will have last coordinate equal to zero, i.e.
\begin{equation}
\bv = \begin{pmatrix} \bw \\0 \end{pmatrix}.
\end{equation}
Assuming that the  calibrated camera is of the form
$P = \begin{bmatrix} R & \bt \end{bmatrix}$,
then \eqref{eq:coneprojection1} will give
\begin{equation}
\bl^T\begin{bmatrix} R & \bt \end{bmatrix} \begin{pmatrix} \bw \\0 \end{pmatrix} = \bl^T R \bw = 0.
\label{eq:cylinderprojection1}
\end{equation}
 The camera geometry for cylinders  is depicted in Fig.~\ref{fig:quadricprojections}.

\section{Triangulation}
We will now describe our triangulation problem. Given a set of $n$ images, with known cameras $P_i$ and with extracted cylinder silhouettes $\bl_i$ we get the following set of equations,
\begin{align}
\bl_i^TP_i\bv & = 0,  \quad i = 1,\ldots,n,
\label{eq:triangulationdir}
\\
\bl_i^TP_iDP_i^T\bl_i &  = 0, \quad i = 1,\ldots,n.
\label{eq:triangulationquad}
\end{align}
In general we have noise in our estimates, so these equations will then not be exactly fulfilled. We could also have gross outliers in our estimated lines, due to e.g. mismatches. In such cases one would like to solve these equations using some robust estimation scheme, such as e.g. RANSAC. Note that both \eqref{eq:triangulationdir} and \eqref{eq:triangulationquad} are linear in the unknown cylinder parameters $\bv$ and $D$, so it would seem that we could use standard linear methods based on e.g. singular value decomposition (SVD). However, as we will show, the problem exhibits a number of complicating factors that make direct linear methods unsuitable. The main issues will be related to \eqref{eq:triangulationquad}. The direction of the cylinder $\bw$ can be linearly estimated from \eqref{eq:triangulationdir}  using either RANSAC based on three-line hypotheses or directly in a least squares manner using SVD of all the $n$ equations. The first problem with directly estimating $D$ linearly using \eqref{eq:triangulationquad} is that we then don't impose the rank deficiency constraint on $D$. We can solve this by aligning the quadric with one of the coordinate axes. If we for instance rotate our current coordinate system so that 
\begin{equation}
R\bw \sim \begin{bmatrix}0 & 1 & 0  \end{bmatrix}^T,
\end{equation}
then the unknown quadric will be on the form
\begin{equation}
D \sim \begin{bmatrix}d_1 & 0 & d_2 & d_3 \\ 0 & 0 & 0 & 0\\
d_2 & 0 & d_4 & d_5\\d_3 & 0 & d_5 & d_6
 \end{bmatrix}.
\end{equation}

\begin{figure}
\begin{center}
  \includegraphics[width=0.46\columnwidth]{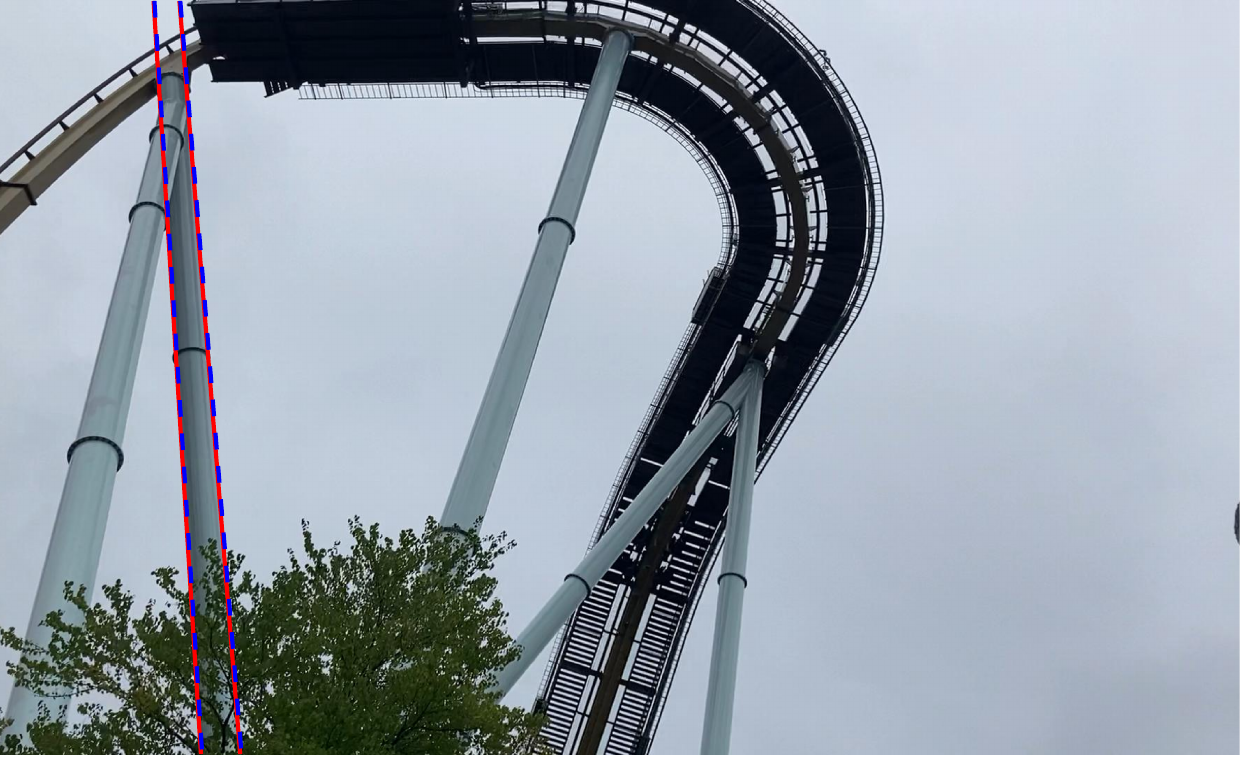}
  \includegraphics[width=0.47\columnwidth]{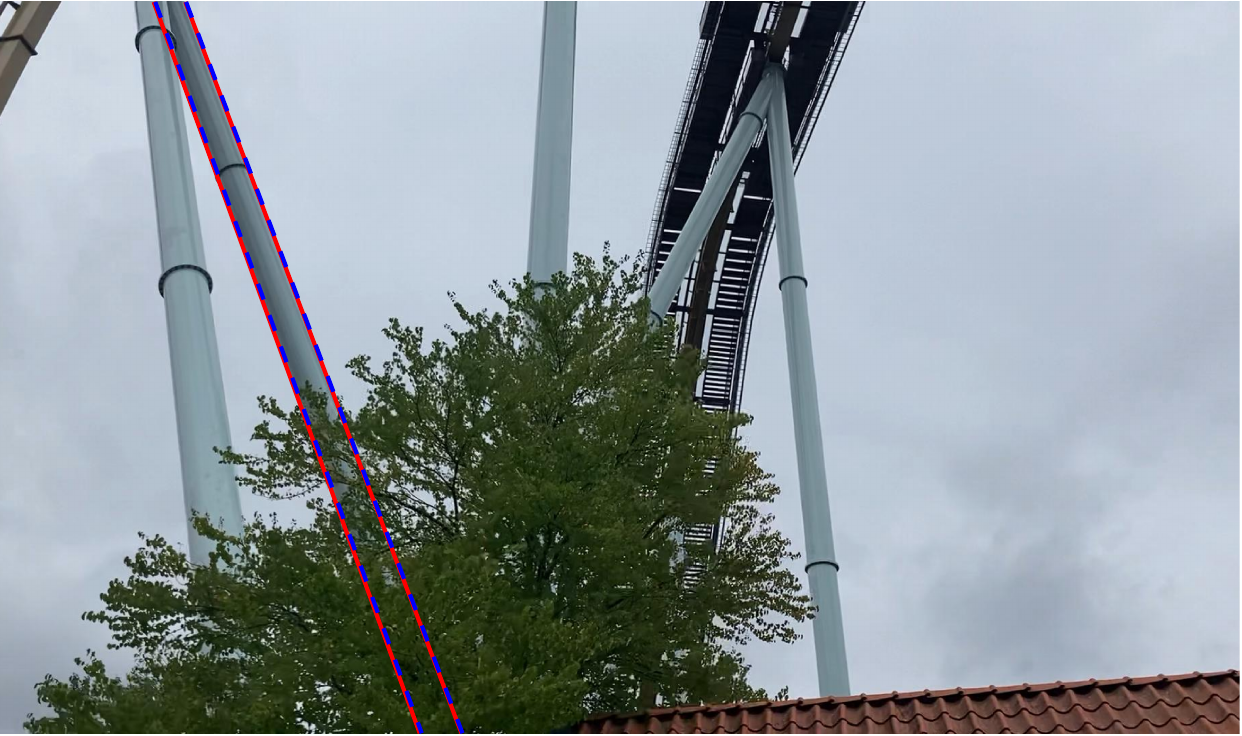}
   \includegraphics[width=0.37\columnwidth]{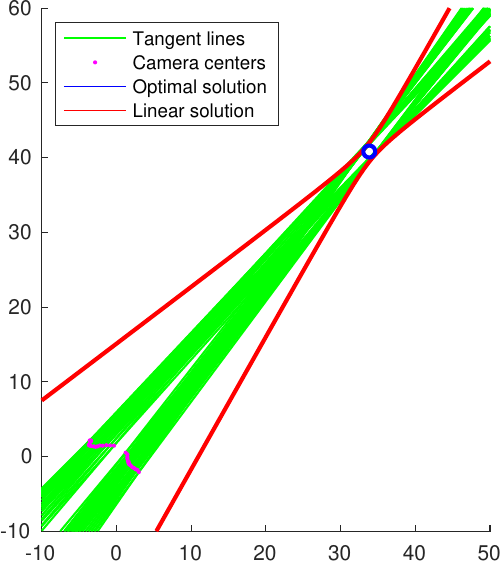}
\includegraphics[width=0.42\columnwidth]{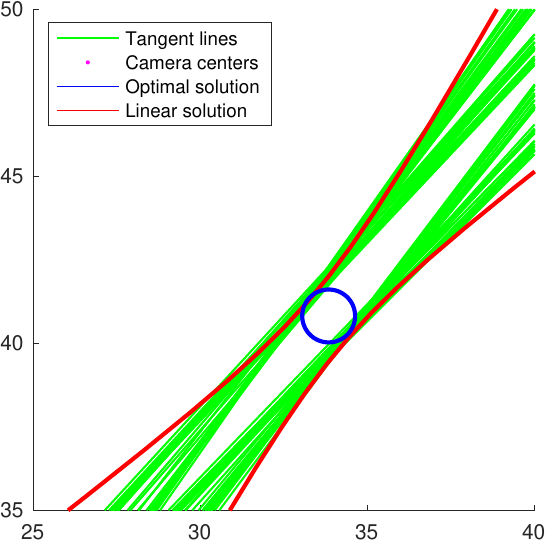}
\end{center}
   \caption{The figure illustrates the problem with estimating the cylinder model using the direct linear method. Top shows two out of 33 images that were used in the experiment. Bottom shows the estimated geometry for one of the cylinders in the images (magnification to the right). Even with a large number of views, the direct linear estimate gives a hyperbola that fits the data very well. This is illustrated in the top views, where this model gives very accurate reprojections (in dashed red). Our proposed constrained least squares solver gives a more reasonable estimate with also accurate reprojections (solid blue in top row). }
\label{fig:badtriang}
\end{figure}

If we also rotate all the cameras $P_i$ using $R$, 
 then the second row of the rotated $P_i$ will only meet zeros in $D$. This enables us to reduce the triangulation problem to a two-dimensional problem. 
\begin{equation}
\br_i^Td\br_i   = 0, \quad i = 1,\ldots,n,
\label{eq:triangulationconic}
\end{equation}
with 
\begin{equation}
d \sim \begin{bmatrix}d_1  & d_2 & d_3 \\
d_2  & d_4 & d_5\\d_3  & d_5 & d_6
 \end{bmatrix},
\label{eq:planarconic}
\end{equation}
where $\br_i$ are the backprojected lines in any plane perpendicular to the $y-$axis.
Here, the problem is now linear  in the unknown planar conic section $d$. If we linearly estimate $d$ from a number of backprojected lines $\br_i$ we should get our sought cylinder. However, it turns out that this problem is inherently badly conditioned. The reason is shown in Figure~\ref{fig:badtriang}. In many cases a cylinder is viewed from a limited  range of angles. This often leads to that there is a hyperbola that is very close to the sought circle or ellipse, in terms of \eqref{eq:triangulationconic}. This means that the linear solution in many such cases will yield an unwanted solution, with very small reprojection error, such as for instance in the example in  Figure~\ref{fig:badtriang}, where the linear solution is compared to one of our proposed solvers. We will in the following describe how we can impose constraints in order to regularize the problem.

\subsection{Dual circle constraints}
We would like to, instead of estimating a general conic section, impose constraint so that our solution is a circle. The cost is that our problem becomes non-linear. The dual representation of a circle at the origin with radius $r$ is given by
\begin{equation}
d_0 \sim \begin{bmatrix}r^2 & 0 & 0 \\ 0 & r^2 &  0\\
0 & 0 & -1
\end{bmatrix}.
\end{equation}
We can find the general expression by transforming this circle with a Euclidean transformation
\begin{equation}
T =  \begin{bmatrix}\cos(\theta) & -\sin(\theta)  & t_x \\ \sin(\theta) & \cos(\theta)  & t_y \\ 0 & 0 &  1
\end{bmatrix}, 
\end{equation}
so that the constrained dual conic $d_c$ is given by
\begin{equation}
d_c \sim Td_0T^T = \begin{bmatrix}r^2- t_x^2&   -t_xt_y &  -t_x\\
-t_xt_y & r^2- t_y^2&  -t_y\\
  -t_x &   -t_y &   -1
\end{bmatrix}.
\label{eq:circleconic}
\end{equation}
Since points are transformed using $T$ as $\tilde{\bf x} = T\bf{x}$, and lines then as $\tilde{\bf r} = T^{-T}{\bf r} \Leftrightarrow {\bf r}  = T^T\tilde{\bf r} $. This gives 
$
{\bf r}^Td_0 {\bf r}= \tilde{\bf r}^TTd_0T^T\tilde{\bf r} = \tilde{\bf r}^Td_c \tilde{\bf r} \Rightarrow d_c = Td_0T^T.
$

This expression can be used to put constraints on the general entries of $d$ in \eqref{eq:planarconic}, since then $\lambda d_c = d$ for some scalar $\lambda \neq 0$. This explicitly  gives 
\begin{align}
\lambda (r^2- t_x^2) = d_1,   \label{eq:1}\\
\lambda (r^2- t_y^2) = d_4, \label{eq:2}\\
\lambda ( -t_xt_y) = d_2, \label{eq:3}\\
\lambda ( -t_x) = d_3, \label{eq:4}\\
\lambda ( -t_y) = d_5, \label{eq:5}\\
\lambda = d_6. \label{eq:6}
\end{align}
Here $\lambda$ is directly given by \eqref{eq:6}. We can then find $t_x$ and $t_y$ from \eqref{eq:4} and \eqref{eq:5}, and $r^2$ from \eqref{eq:1}. Inserting these expressions in \eqref{eq:2}  and \eqref{eq:3} (and multiplying with $d_6$), gives two second degree polynomials that the entries of $d$ need to fulfill. The conics that are circles must then lie on the non-linear manifold defined by 
\begin{equation}
\mathcal{M}_c = \{ d | d_3d_5-d_2d_6 = 0, d_5^2-d_3^2+d_1d_6-d_4d_6 = 0\}.
\end{equation}
We can now define our constrained triangulation in the following way
\begin{equation}
\br_i^Td\br_i   = 0, \quad i = 1,\ldots,n, \quad d \in \mathcal{M}_c.
\label{eq:triangulationcircle}
\end{equation}
\begin{remark}
We could of course also choose to work directly with the parametrization given by \eqref{eq:circleconic} when we develop our solvers and methods. However, often it's much more numerically stable to work with a manifold and use the linear constraints from the original equations to parametrize using the nullspace (compare using the trace-constraints on the essential matrix to construct stable minimal solvers).
\end{remark}
\subsection{Minimal three line solver}
\label{sec:minimal}
The manifold defined by ${M}_c$ has three degrees of freedom, since $d$ has six parameters defined up to scale, and we have two non-linear constraints. The degrees of freedom corresponds to the unknown $r$,$t_x$ and $t_y$. This means that we should be able to recover $d$ minimally from three constraints. Since each silhouette line will give an equation according to \eqref{eq:triangulationcircle} we need three lines to determine $d$. We use the linear constraints to parametrize $d$, with two unknowns $\alpha$ and $\beta$, as
\begin{equation}
d = \alpha d_\alpha +  \beta d_\beta +  d_\gamma,
\end{equation}
where $d_\alpha$, $d_\beta$ and $d_\gamma$ only depend on data.
Inserting this parametrization of $d$ in the constraints of $\mathcal{M}_c $ gives two second degree polynomials in $\alpha$ and $\beta$. The problem has in general four solutions, and one can solve the system of equations in a number of ways, e.g. using resultant techniques \cite{cox2013ideals}. We have constructed a fast solver (using the automatic generator from \cite{larsson2017efficient}), that runs in $7 \mu s$ on a standard laptop (Macbook pro 2.5 GHz Dual-Core Intel Core i7 running Matlab 2020b), based on a Matlab/Mex-C++ implementation. 
\subsection{Least squares solver}
\label{sec:optimal}
If we have no outliers in the data, or have eliminated them in some way, it would be desirable to directly find the solution using all the available equations. One way to solve this would be to find the least squares solution that also adheres to the non-linear constraints on $d$, i.e. find the solution to 
\begin{equation}
\min_{d \in \mathcal{M}_c} \sum_{i =1}^n (\br_i^Td\br_i)^2.
\label{eq:ls_circle}
\end{equation}
Here we need to also fix the scale in some way, since otherwise the trivial (and non-valid general conic matrix) $d = \bf{0}$ would attain the minimum. In order to avoid this we set $d_6 = -1$. 
One way to find the solution to this  is to find all local minima, and choose the best one among them. To this end we  formulate the Lagrangian
\begin{align}
\begin{split}
L(d,\lambda_1,\lambda_2)  = \sum_{i =1}^n (\br_i^Td\br_i)^2 + \lambda_1(d_3d_5-d_2d_6 ) + \\  \lambda_2(d_5^2-d_3^2+d_1d_6-d_4d_6), 
\end{split}
\end{align}
and then the local minima are given by solving $\nabla L = \bf{0}$.
This gives seven equations in the seven unknowns $(d_1,\ldots,d_5,\lambda_1,\lambda_2)$.  The four equations 
\begin{equation}
(\frac{\partial L}{\partial d_1} = 0,\frac{\partial L}{\partial d_2} = 0,\frac{\partial L}{\partial \lambda_1} = 0,\frac{\partial L}{\partial \lambda_2} = 0),
\end{equation}
will be linear in $(d_1,d_2, \lambda_1, \lambda_2)$, and we can hence use these equations to linearly eliminate them. Inserting the expressions for $(d_1,d_2, \lambda_1, \lambda_2)$ in the two remaining equations 
\begin{equation}
(\frac{\partial L}{\partial d_3} = 0,\frac{\partial L}{\partial d_4} = 0,\frac{\partial L}{\partial d_5} = 0),
\end{equation}
gives three polynomials in the remaining three unknown $(d_3,d_4,d_5)$ (two with total degree three, and one with total degree two). This system has in general up to nine solutions. We have again used the automatic generator from \cite{larsson2017efficient} to construct a solver. The resulting solver has an elimination template of size $25 \times 34$ and runs in $30 \mu s$ on a standard laptop (Macbook pro 2.5 GHz Dual-Core Intel Core i7 running Matlab 2020b), based on a Matlab/Mex-C++ implementation. 
The solution to \eqref{eq:ls_circle} is found by evaluating the found local minima and choosing the one that gives the smallest total error. 

\subsection{Robust and accurate triangulation}
The results in the previous sections can be used to accurately and robustly triangulate cylinder surfaces. The basic steps are summarized in Algorithm~\ref{alg:system_proposed}. Given a number of silhouette lines we linearly estimate the cylinder direction, and then we use our proposed solvers to estimate the circular cross section. Our estimates are based on algebraic costs, and one can refine these estimates using standard optimization methods, based on reprojection errors. However, from our experiments, the proposed methods seem to directly give very accurate results. 
\begin{algorithm}
\caption{Proposed Triangulation System (main contributions in bold)}
\begin{algorithmic}[1]
	\Require A set of measured image silhouette lines, with corresponding camera matrices. 
	\State Find the cylinder direction $\bw$, linearly using \eqref{eq:triangulationdir}, either using all lines in a least squares sense or robustly using RANSAC.
	\State Find a rotation that moves $\bw$ to $\begin{bmatrix}0 & 1 & 0  \end{bmatrix}^T$.
	\State {\bf Use the rotation to transform the problem to the constrained form \eqref{eq:triangulationcircle}.}
	\State {\bf Solve for the unknown $d \in \mathcal{M}_c$ robustly using RANSAC (Section ~\ref{sec:minimal}) or in a least squares sense (Section ~\ref{sec:optimal}) }
	\State Transform back to the original coordinate system.
	\State Use non-linear refinement of all unknowns, e.g., by using gradient descent or Levenberg-Marquardt  (optional, all results in the paper are without the use of refinement).
\end{algorithmic}
\label{alg:system_proposed}
\end{algorithm}

\section{Experimental evaluation}

We will in this section evaluate our proposed solvers. We test the basic solvers numerically on synthetic data and evaluate how they can be used efficiently to triangulate cylinders in a number of different real scenarios. 
\subsection{The problem using the direct linear method}
We will start by an example where the direct linear method fails to estimate an accurate cylinder model. 
We acquired 33 images of a roller-coaster with a number of steel pillars, that can be modelled accurately using cylinders. Two examples of the images are shown in the top row of Figure~\ref{fig:badtriang}. We used an automatic SfM estimation system (COLMAP \cite{schoenberger2016sfm,schoenberger2016mvs}) to estimate the camera geometry. 
We then manually extracted silhouette lines for one of the pillars in all the images. Running the direct linear method using all the images will in this case lead to a hyperbolic quadric instead of a cylinder. The reason can be seen in the close-up of the two-dimensional geometry shown in the bottom right of the figure. The tangent lines will be very close to both a circle and and hyperbola. This means that both models will give small errors and accurate reprojections, as can be seen in the top of the figure. The reprojection silhouette lines are shown for both the linear estimate (dashed red) and the proposed least squares solver (solid blue). This means that if we want to find the true model, we need to put more constraints on it, and not  simply state that is should be a quadric surface. The problem typically appears when we do not have images from all view-points---A scenario that is common in most real world applications.  

\subsection{Synthetic numerical evaluation of solvers}
\begin{figure*}
\begin{center}

\includegraphics[width=0.48\columnwidth]{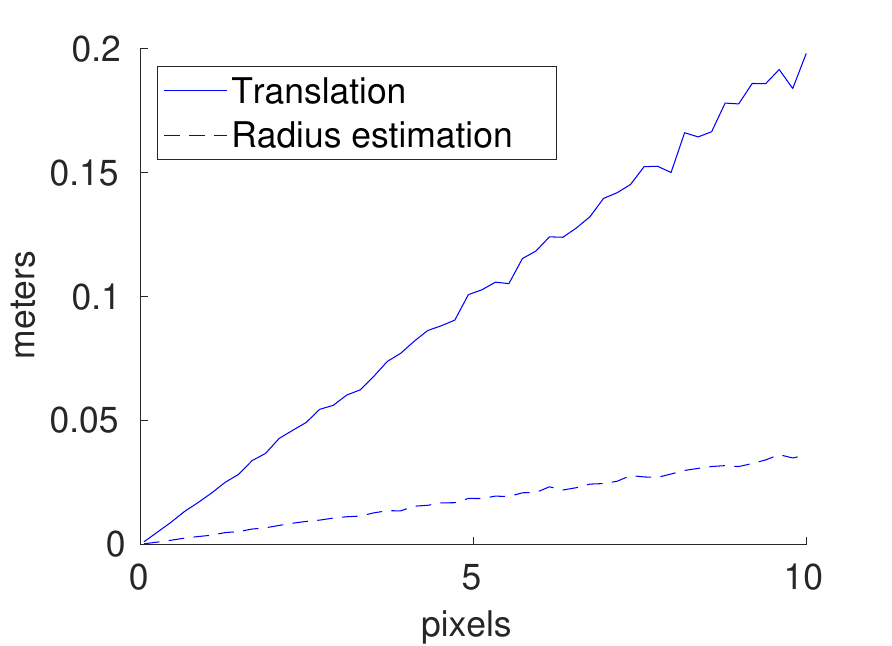} 
\includegraphics[width=0.48\columnwidth]{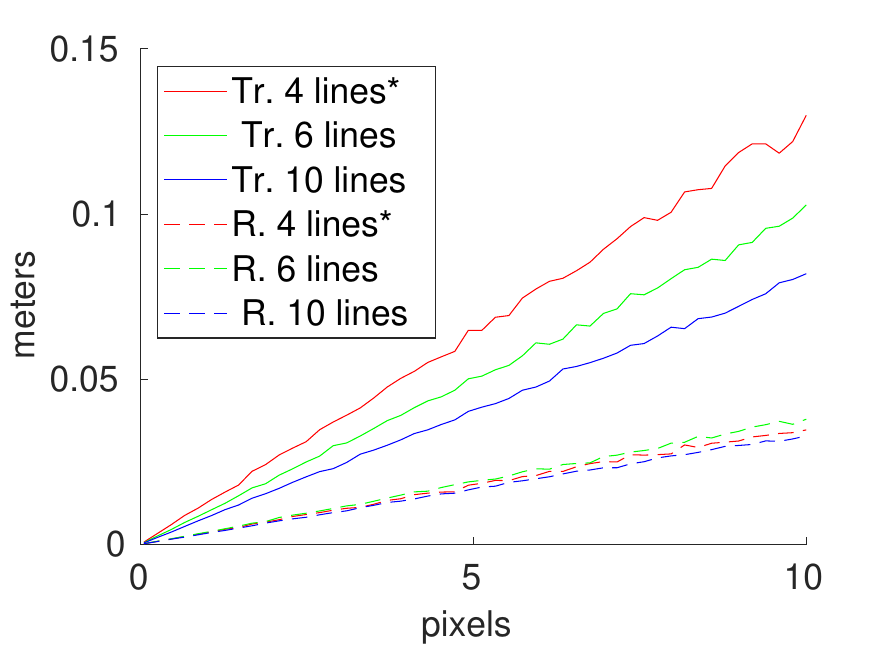} 
\includegraphics[width=0.48\columnwidth]{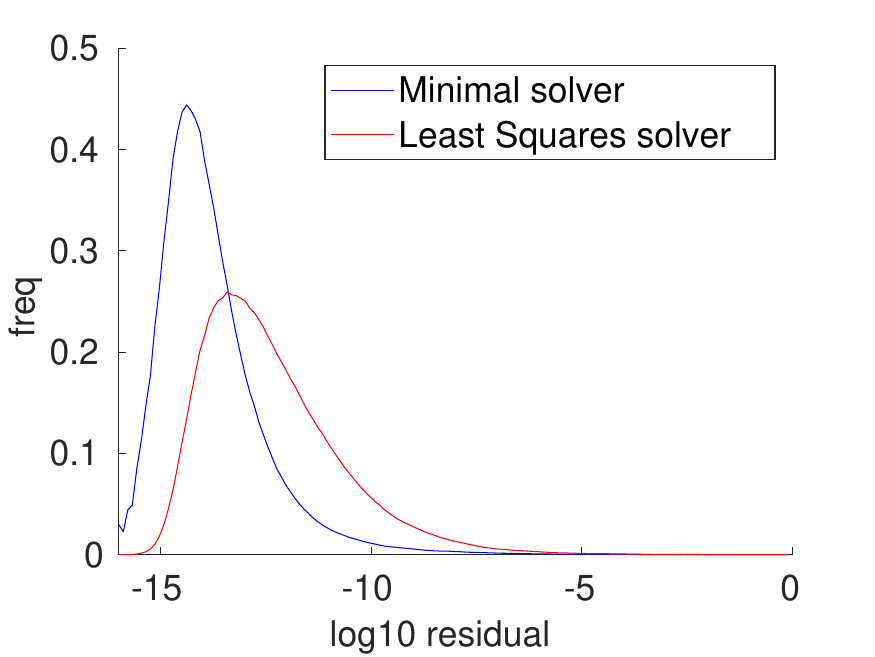}
\end{center}
   \caption{ Left: Translation and radius estimation error when using the minimal solver. Right: Translation and radius estimation error using the least squares solver and four,  six and ten lines (two, three or five cameras). In the case of four lines the best fitting solution has been chosen. Bottom: Distribution of residuals from the two solvers. }
\label{fig:synth}
\end{figure*}

We perform a number of synthetic tests to show the robustness and accuracy of our solvers. Firstly numerics without noise,  and secondly accuracy in the presence of noise of varying degree. The numerical stability was tested by calculating the maximum absolute value of the residuals, without noise in the measuremenst. Results are shown in the bottom panel of Figure~\ref{fig:synth}, where we get close to machine precision.
Robustness to noise was tested on an already rectified system, meaning that the image lines are reduced to  the form $x = a$. Noise was introduced to the lines as
\begin{equation}
x = a + N(\sigma,0).
\end{equation}
Since the problem is scalable we need to mention our experimental setup. We put the cylinder in origo with a random radius $0.5 < r < 2$ and the cameras sampled uniformly random in  the $40 \times40$ square centered around origo, omitting the $6\times6$ square in the middle. The standard deviation was tested in the range $0<\sigma<0.02$, which corresponds to [0 10] pixels for a camera with focal length $f=500$. 
The results for the minimal solver is shown in the left panel of Figure \ref{fig:synth}. The minimal solver produces a maximum of four real solutions. Without use of extra lines or knowledge of the setup it is not possible to know which solution is the correct one. We therefore pick the one closest to the true value. The least square solver has one more parameter to investigate, how many cameras or lines to include.  The results for different number of lines, are shown to the right in Figure \ref{fig:synth}. Errors are measured in Euclidian distance. For the four-line case we are close to the minimal case, and there are often several local minima that have small total error. For this reason we report the solution closest to the gound truth cylinder. For the six and ten line we report the global optimum. 
 Choosing the correct solution could be aided for example by only considering solutions in front of the camera or evaluating using more lines. Unsurprisingly the estimations get better with more samples. Note that the error response to noise is close to linear.





\subsection{Comparison to other methods}
In this section we will compare our least squares solver to the direct linear method and the cylinder triangulation method of Navab et al.\ \cite{navab2006canonical}, in a controlled real experiment. We took a number of images (18 in total) of an object with known geometry (a soda can). The setup is depicted in the supplemental material. We registered the cameras and extracted silhouette lines from all images. We then selected a number of views randomly and ran the different triangulation methods. All three methods use the same way of finding the cylinder direction, so we only compare the estimated cross sections to the ground truth. Table~\ref{t_comparison} shows the mean Frobenius norm  between the ground truth conic and the respective estimated conics, based on 10~000 runs for each  different number of used views.  One can see that the linear method suffers from the degeneracy for low number of views. The method of Navab et al.\ doesn't use the data in an optimal way, and gives quite innacurate results even for large number of views. 
\begin{table}
\centering
\caption{Mean Frobenius norms of differences between estimated conic cross sections and ground truth, for different triangulation methods. Best results in bold for each number of used images. } 
\label{t_comparison}
\begin{tabular}{@{}cccc@{}} \toprule
Number of Images & Proposed & Linear & Navab et al. \cite{navab2006canonical}\\
\midrule 
 2 & 1.447  & 50.53  &  \bf{0.885}\\
 3 &  \bf{0.497}  &  0.556  &  0.512\\
 5 &   \bf{0.294} &   0.299  &  0.389\\
 10 &  \bf{0.148} &  0.151 &   0.302\\
  15 &  \bf{0.072}  &  0.074  &  0.278\\
\bottomrule
\end{tabular}
\end{table}

\subsection{Robust matching using exhaustive sampling}
\begin{figure}
\begin{center}
   \includegraphics[width=0.32\columnwidth]{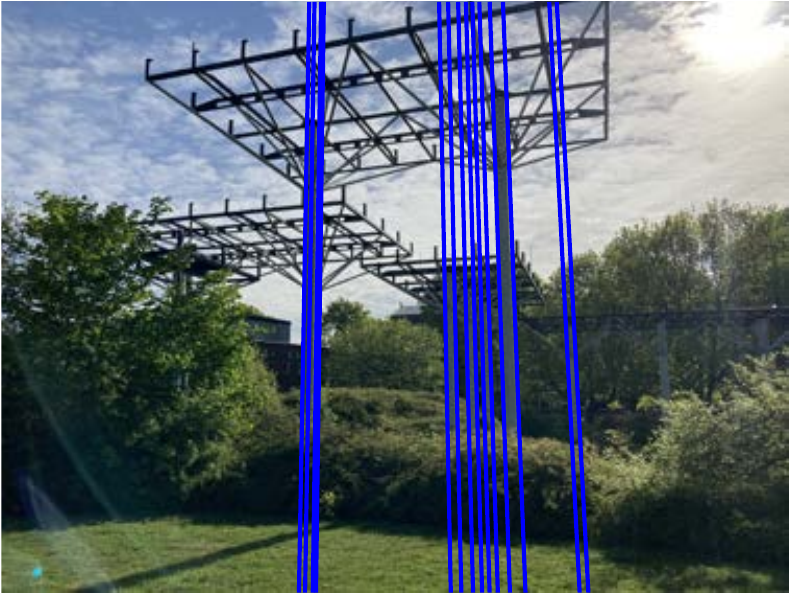}
   \includegraphics[width=0.32\columnwidth]{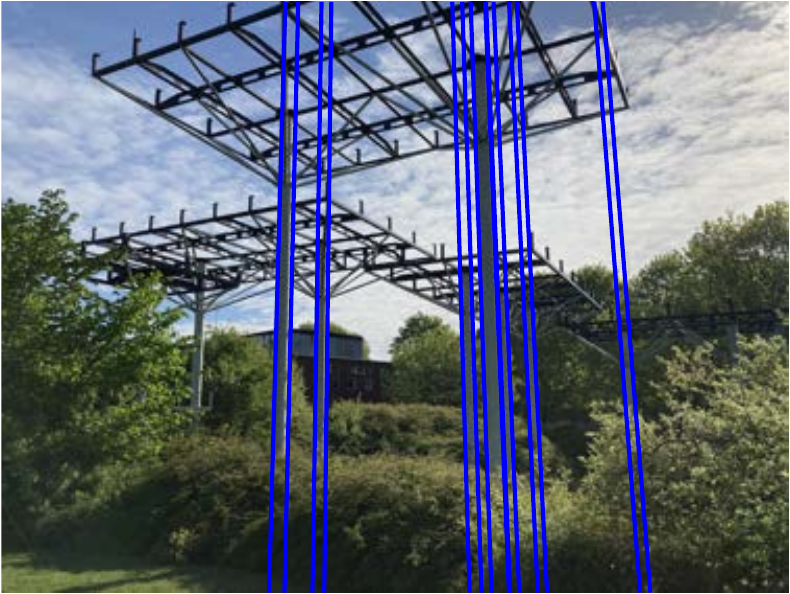}
   \includegraphics[width=0.32\columnwidth]{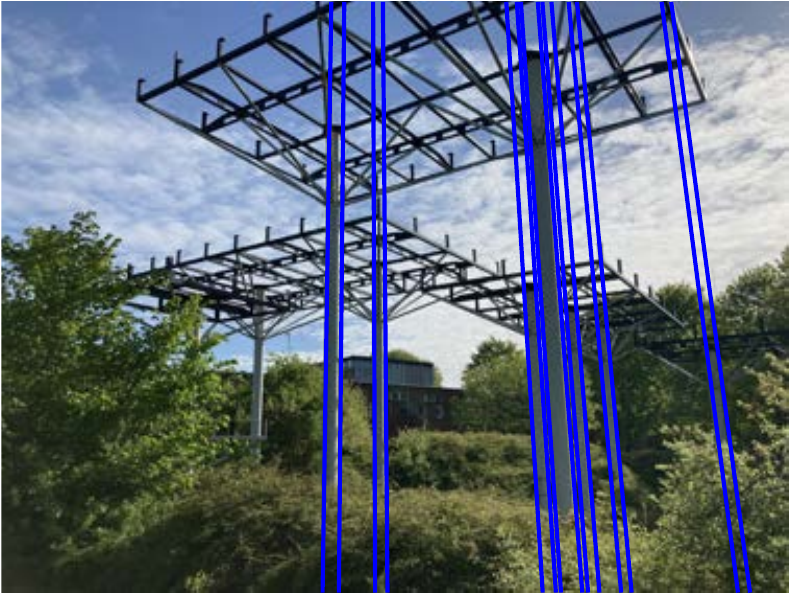}
   \includegraphics[width=0.49\columnwidth]{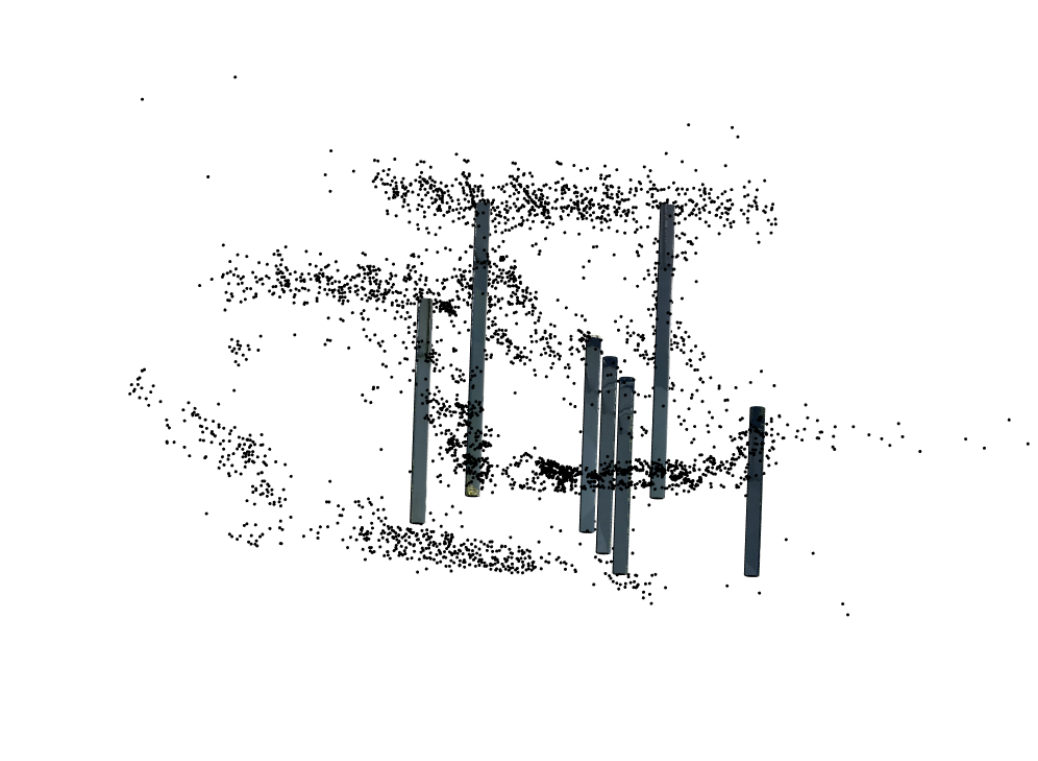}
   \includegraphics[width=0.49\columnwidth]{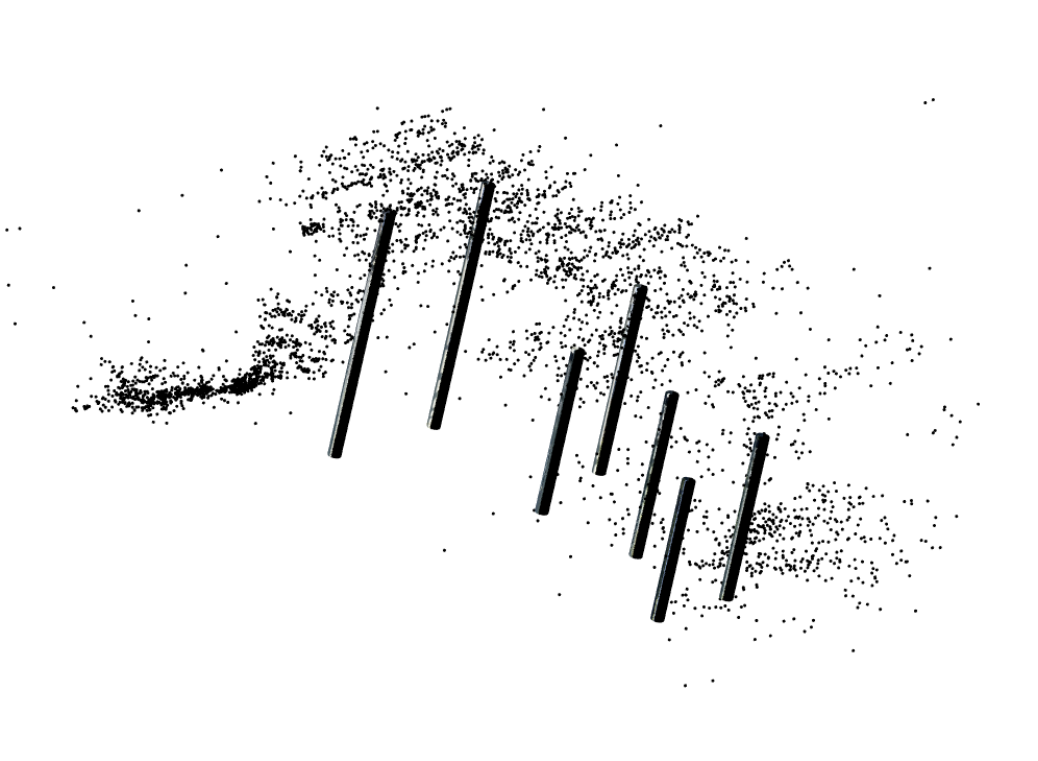}
 \end{center}
   \caption{The figure shows the result of estimating several similar cylinders, where we have extracted silhouette lines in several images, but do not have image line matches available across views. Top row shows reprojected silhouette lines in blue for three of the twelve images used in the experiment. Bottom shows two views of the seven  reconstructed  cylinders, together with the sparse points used for registering the cameras.}
\label{fig:fontana3}	
\end{figure}

In many cases one has a scene with several cylinders with similar appearance in a scene. Examples are the pylons in Figures~\ref{fig:badtriang} and \ref{fig:fontana3}. Here, a natural scenario is that we have extracted silhoutte lines in a number of images, but we do not have matching information between them. We will now test how our minimal solver can be used to efficiently and robustly find the matchings at the same time as we estimate the cylinder models. We have manually extracted silhouette lines from a set of twelve images of a water fountain. Three of the images are shown in Figure~\ref{fig:fontana3}. We have used COLMAP \cite{schoenberger2016sfm,schoenberger2016mvs} to estimate the camera geometry.
 Here we also assume that in the first image, for the  cylinders that we have detected, both the left and the right silhoutte line have been found. In this case we know that all cylinders are parallell, so we can estimate a common vanishing direction for all cylinders in all images jointly. 
Given that we only need three lines to minimally estimate the circular cross section of one of the cylinders, it's possible to do exhaustive sampling in linear complexity in the number of extracted silhouette lines. For each pair of image lines in the first image, we sample one more line from one of the other images, and estimate a cylinder, using our proposed minimal solver from Section~\ref{sec:minimal}. For each such model we test how many of all possible lines fit the model. After testing all possible lines, we choose the model with most inliers. This gives a model and matching lines for the chosen cylinder in the first image. We then repeat this for each silhouette pair in image one. The resulting set of cylinder models is shown in the bottom of Figure~\ref{fig:fontana3}, together with the sparse points estimated by COLMAP. The top of the figure shows the reprojected cylinder silhouettes (in blue) in three of the images. We see that it's possible in this case to do the matching and estimation in parallell for a number of images and cylinders. Note here that since we do not have explicit correspondences we have an extremely high outlier rate. For the seven reconstructed cylinders the average outlier rate was in this case $89\%$.

\subsection{Using only one silhouette line in each image}
\begin{figure*}
\begin{center}
  \includegraphics[width=0.25\columnwidth]{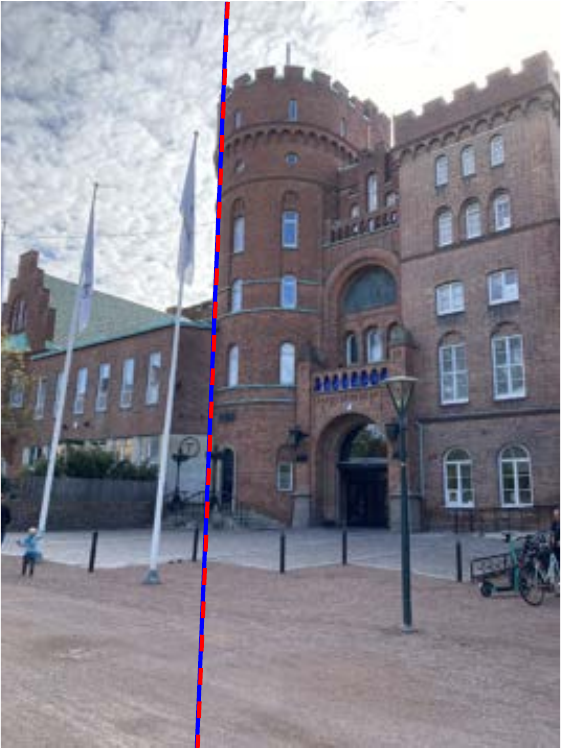}
  \includegraphics[width=0.25\columnwidth]{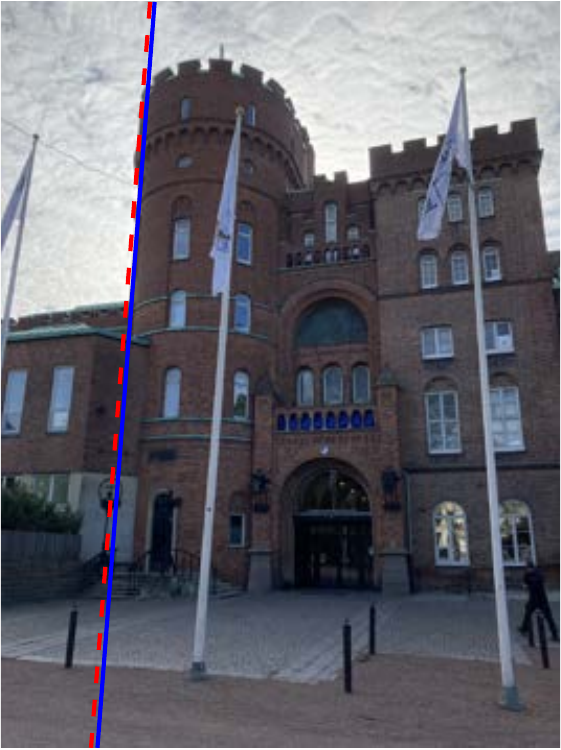}
  \includegraphics[width=0.25\columnwidth]{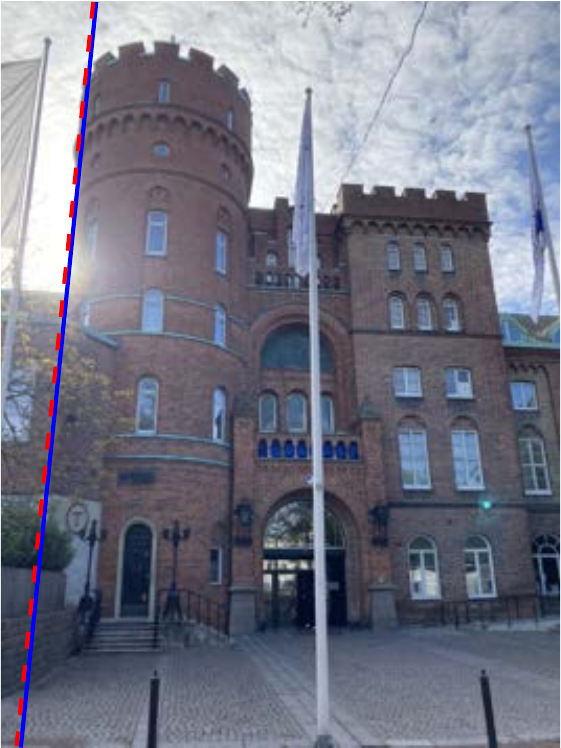}
  \includegraphics[width=0.25\columnwidth]{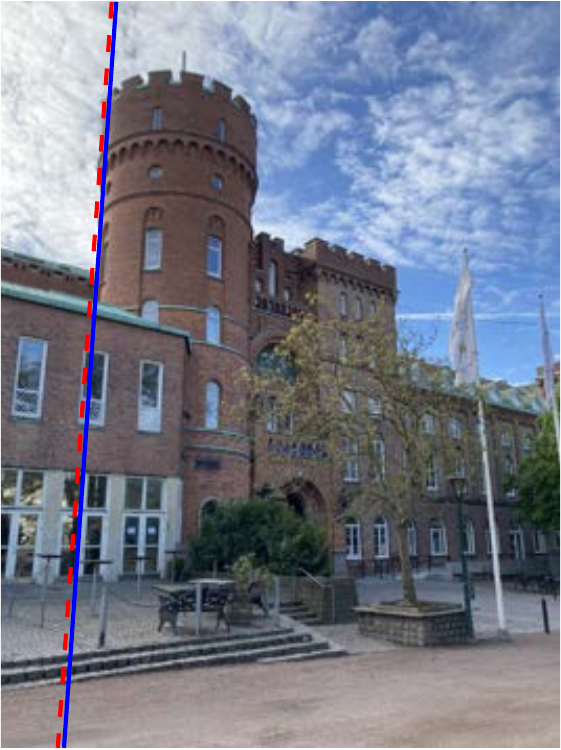}
  \includegraphics[width=0.25\columnwidth]{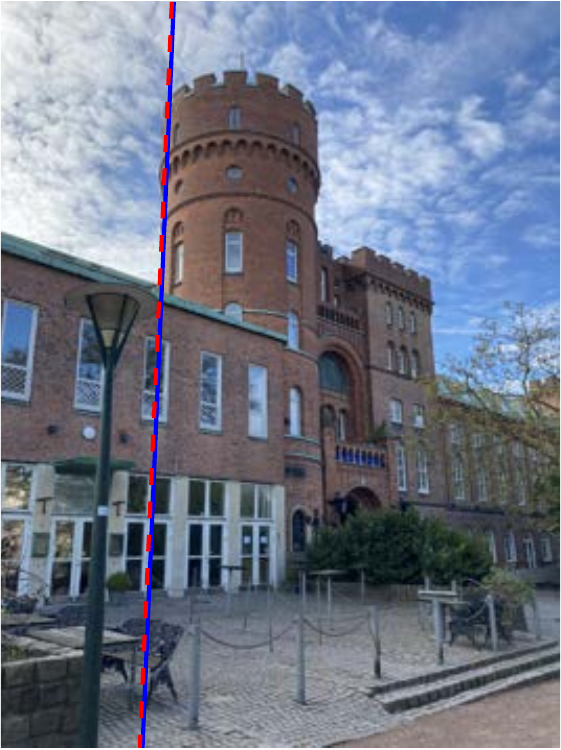}
  \includegraphics[width=0.25\columnwidth]{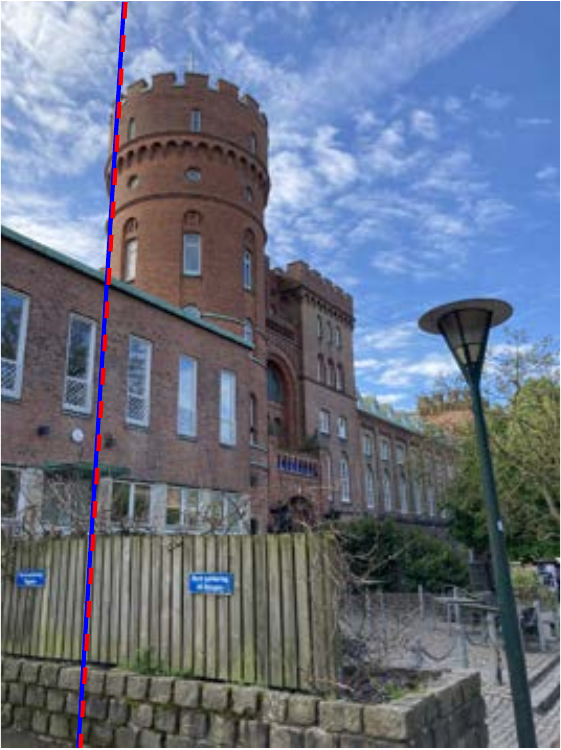}
   \includegraphics[width=0.42\columnwidth]{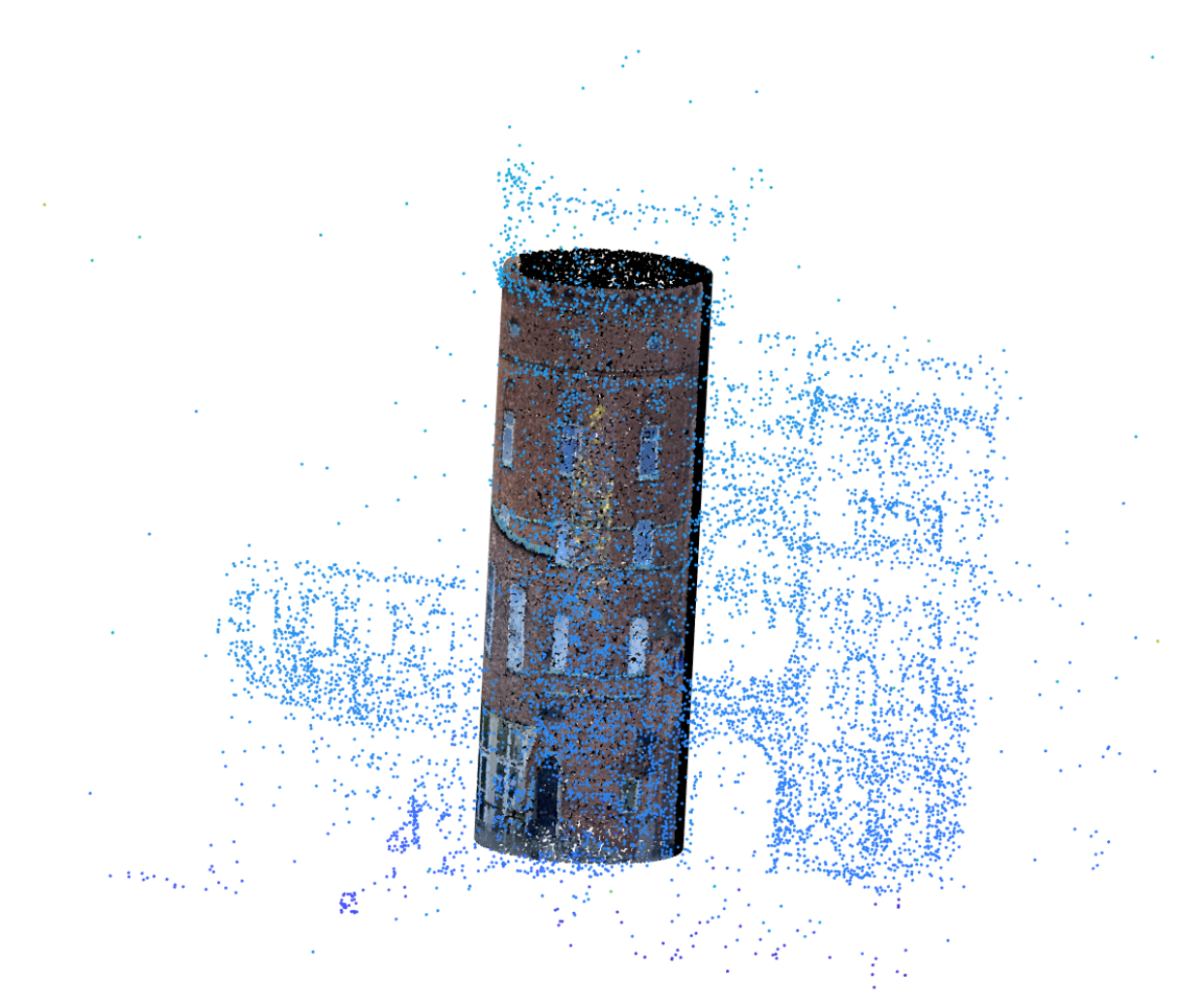}
   \includegraphics[width=0.42\columnwidth]{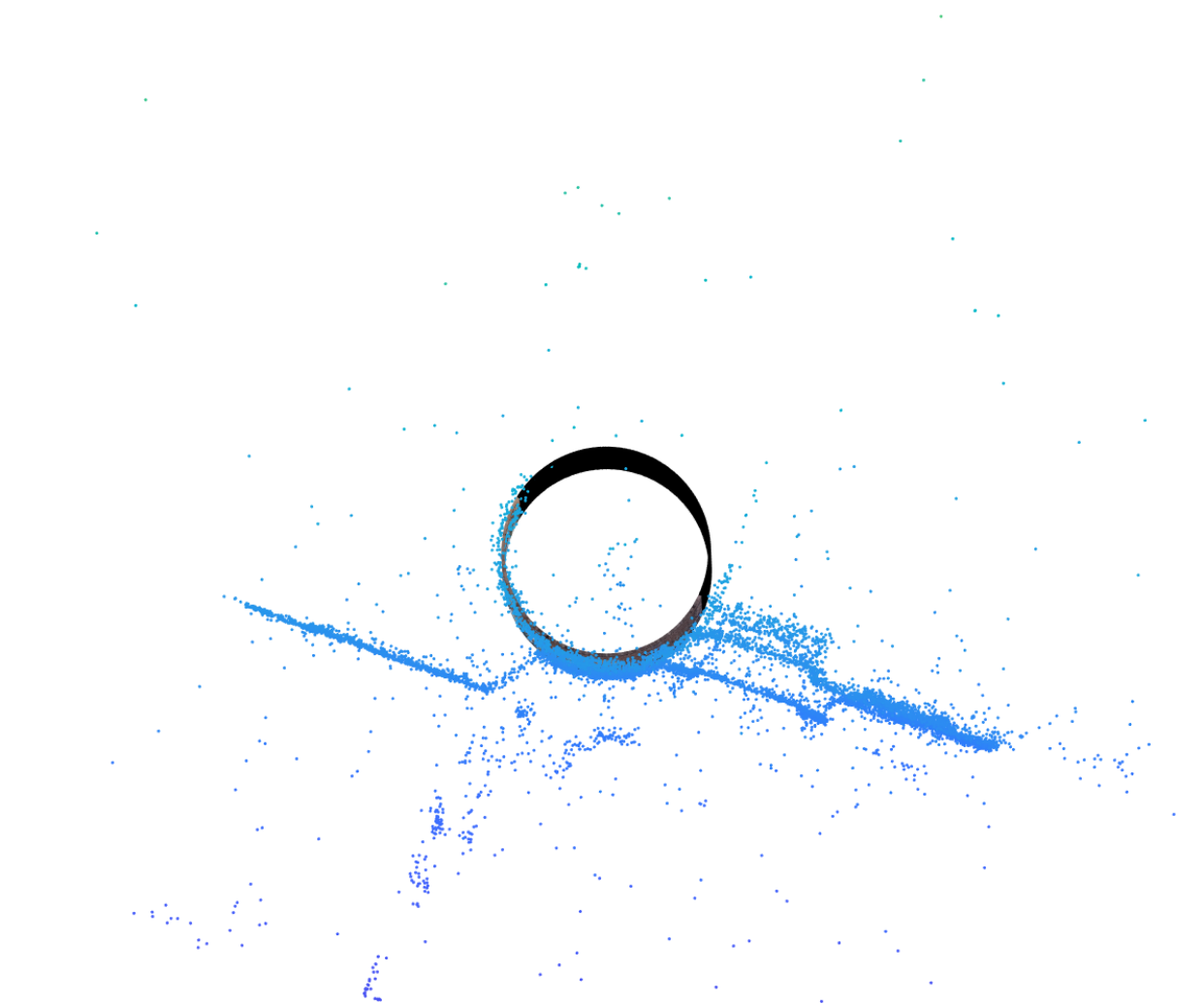}
   \includegraphics[width=0.42\columnwidth]{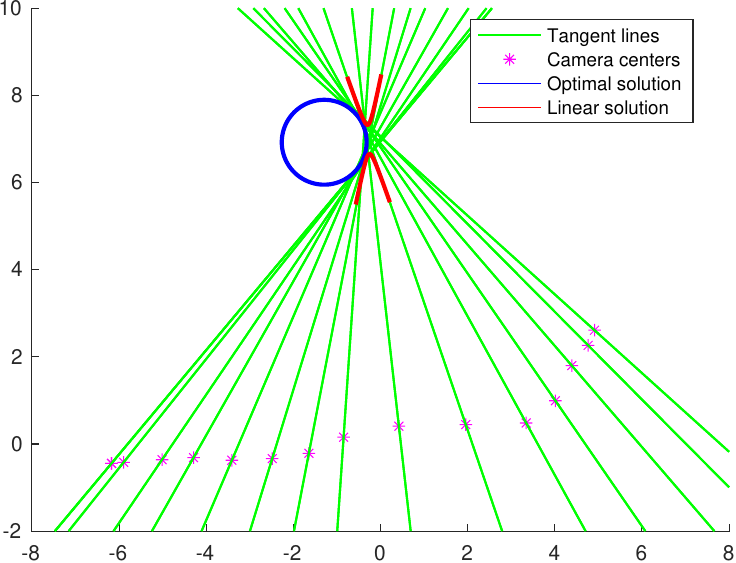}
\end{center}
   \caption{An example where we only use one of the silhouette lines in each image. Top row shows six out of fifteen images, together with reprojected lines for both the proposed constrained least squares solver (solid blue) and direct linear solution (dashed red). Bottom row shows two views of the estimated cylinder model using the proposed least squares solver. Bottom right shows the two-dimensional view of the geometry with both the estimated models, where the direct linear solver fails to produce a reasonable solution.}
\label{fig:afborgen}
\end{figure*}

One case where the linear estimation of the conic cross-section of a cylinder typically gives poor results is when we only have one of the silhouette lines available in each image. This can occur in many cases due to e.g. occlusion or when the actual cylinder is not a full cylinder. To test how well our method fares in this case we took a set of fifteen images of a tower building. A number of the images are shown in the top of Figure~\ref{fig:afborgen}. We manually extracted only the left silhouette line in each image, and again registered the cameras using COLMAP \cite{schoenberger2016sfm,schoenberger2016mvs}. We then estimated the cylinder both using our constrained least squares solver and the general linear solver. Even though we have many images, taken from quite varying viewing angles, the linear solver fails to give a resonable solution, as can be seen in the lower right panel of Figure~\ref{fig:afborgen}. 
The proposed constrained least squares solver, on the other hand, gives a reasonable estimate of the cylinder, that also aligns well with the sparse points estimated by COLMAP. The resulting reconstruction is shown in the bottom of Figure~\ref{fig:afborgen}. The two-dimensional cross-section estimates (seen from below) are shown in the lower right in the figure, together with the camera centers and back-projected silhouette lines. The top of the figure shows six frames from the sequence together with reprojected silhouette lines (proposed solver in solid blue and linear estimate in dashed red). One can see that the linear estimate also gives very accurate reprojection silhouettes, so the problem is inherently ill-posed if we do not constrain our cylinder model. 
For more examples see supplementary material.

\section{Conclusion}
We have in this paper investigated how infinite cylinders can be reconstructed in 3D by triangulation of silhouette lines in a number of images with known camera geometry. It turns out that in many real scenarios the linear estimate of a cylinder will be ill-posed, and we propose to constrain the estimates to a manifold described by two second degree polynomial constraints. Using these constraints, we derived two novel solvers for robust and least squares estimation respectively. The solvers have been tested on real and synthetic data, and shown to give robust and accurate results, compared to previous methods.
We have in this paper targeted the purely geometric estimation problem, and of course much work remains if one wants to use such methods in real systems. Components for silhouette feature extraction and  tentative matching are needed, as well as non-linear refinement of final estimates. Other geometric estimates, e.g. for relative pose and full SfM would also be interesting to further investigate. We also believe that these cylindric representation can be very useful for further processing tasks, involving more semantic analysis, such as recognition and scene analysis. 


\newpage
\bibliographystyle{splncs04}
\bibliography{ref}

\end{document}